# Learning Hidden Markov Models for Regression using Path Aggregation

**Keith Noto**
Dept. of Computer Science and Engineering
University of California San Diego
La Jolla, CA 92093
knoto@cs.ucsd.edu

**Mark Craven**
Dept. of Biostatistics and Medical Informatics
University of Wisconsin
Madison, WI 53706
craven@biostat.wisc.edu

## Abstract

We consider the task of learning mappings from sequential data to real-valued responses. We present and evaluate an approach to learning a type of hidden Markov model (HMM) for regression. The learning process involves inferring the structure and parameters of a conventional HMM, while simultaneously learning a regression model that maps features that characterize paths through the model to continuous responses. Our results, in both synthetic and biological domains, demonstrate the value of jointly learning the two components of our approach.

## 1 Introduction

A wide array of problems in speech and language processing, biology, vision, and other application domains involve learning models that map sequences of inputs into some type of output. Common types of task include learning models that classify sequences (*e.g.*, [12]), segment or parse them (*e.g.*, [13]), or map input sequences to output sequences (*e.g.*, [3]). Here we consider the task of learning models that map input sequences to real-valued responses. We present an approach to this problem that involves simultaneously learning a hidden Markov model (HMM) and a function that maps paths through this model to real-valued responses. We evaluate our approach using synthetic data sets and a large collection from a yeast genomics study.

The type of task that we consider is illustrated in Figure 1. This is a type of regression task in that the learner must induce a mapping from a given input sequence to a real-valued response. In particular, we assume that the real-valued responses can be represented as a function of the presence and arrangement of particular *motifs* that occur in the sequences. Each of these motifs is a pattern that allows some variability in the subsequences that match it.

We assume that neither these motifs nor their locations in the sequences are given to the learner, but instead must be discovered during the learning process. The learner must also determine the extent to which each motif and its relationships to other motifs contribute to the response variable.

Our research is motivated by a class of problems in computational biology that involve inferring the extent to which particular properties of genomic sequences determine certain responses in a cell. For example, the level at which an individual gene is expressed in a given condition often depends on the presence of particular *activators* that bind to sequence motifs nearby the gene. Moreover, the number of binding motifs, their arrangement in the sequence, and intrinsic properties of the motifs themselves may contribute to the response level of the gene. Thus, in order to explain the expression levels of genes in some condition, a model needs to be able to map these sequence properties into continuous values.

The approach that we present involves simultaneously learning (i) the structure and parameters of a hidden Markov model, and (ii) a function that maps paths through the model to real-valued responses. The hidden Markov model is able to represent the relevant sequence motifs and the regression model is able to represent the mapping from occurrences of these motifs to the response variable.

There are several bodies of related research. First, there is a wide variety of architectures and methods for learning HMMs [19], stochastic context-free grammars [17], and related probabilistic sequence models, such as conditional random fields [13]. For some types of problems, these models include continuous random variables. Typically these continuous variables depend only on a few other variables, and the dependencies are encoded at the outset. In our models, in contrast, the continuous response variable may depend on quite a few variables that characterize the input sequence, and these variables and their dependencies are determined during the learning process.

| | | |
|---|---|---|
| 1 | ...acc**ccgatgag**aaaatt**ccgatgag**tttagaaggta... | 6.0 |
| 2 | ...aaagaaaaaaaaaaaaagaaaagaaaaaa**gagatgag**a... | 3.2 |
| 3 | ...caaagg**gcgatgag**ataaaagcgat**aaaaattttt**gag... | 9.1 |
| 4 | ...ggcgcgactcg**gcgatgag**atgagcagagataaagaca... | 3.1 |
| 5 | ...gagctggatgct**ataaatttctt**aggtataaagtacga... | 5.9 |
| 6 | ...aac**aagaatttatt**catcattaa**gcgatgcg**actcact... | 8.9 |
| 7 | ...ccattttttctctcttttataagaata**aaatgttttt**a... | 6.1 |
| 8 | ...**gcgatgag**tttcactaaaa**ataaattttc**ttttccaa... | 9.2 |

Figure 1: An example of the sequence-based regression task that we are addressing. Each row in the figure represents a particular training instance. Each instance consists of a DNA sequence and an associated real-valued output. The sequences in this example contain two types of motifs; $m_1$ whose consensus sequence is gcgatgag and $m_2$ whose consensus sequence is aaaaattttt. In the tasks we consider, the motifs and their occurrences are hidden. The learning task involves discovering the motifs and their occurrences in the sequences, and inferring a function that maps motif occurrences to the real value associated with each sequence. In this example, $y \approx 3 \times v_1 + 6 \times v_2$, where $v_1$ represents the number of occurrences of $m_1$ and $v_2$ represents the number of occurrences of $m_2$.

There is also large corpus of work on the topic of regression methods [7]. Most regression methods assume that each instance is represented using a fixed-size set of pre-defined variables. Our approach, on the other hand, assumes that each instance is represented by a sequence of values, but these sequences may vary in their lengths and the positions of the relevant sequence elements may vary as well. Moreover, our method is designed to derive a set of variables, from the given sequences, that are predictive of the response variable.

Various groups have devised kernels defined over sequences that provide mappings from sequence features to real numbers. These string kernels can be used to map sequences to feature vectors which can then be used for regression or classification [15]. However, these kernels encode predefined sequence features. In contrast, our method is designed to learn which sequence features best provide input to the regression part of the model. Jaakkola et al. [8] have used HMMs to identify the relevant aspects of sequences then, in a second step, the Fisher kernel for classification based on the HMM representations. Our experiments in Section 3 indicate that more accurate models can be learned by using the training signal to guide the discovery of relevant sequence features.

Several inductive logic programming (ILP) methods for learning regression models have been previously developed [9, 11]. The algorithms are similar to ours in that they can handle variable-sized descriptions of in-stances and they employ an expressive representation for regression tasks. They differ from our approach in that they are not designed to discover sequence motifs and use properties of these motifs in the regression model. This aspect of our approach is essential for the problems we consider.

A variety of methods have been developed for discovering motifs in biological sequences [1, 14, 16], and for identifying arrangements of motifs that are involved in particular biological processes [20, 22]. These methods are designed for either unsupervised pattern discovery or supervised classification tasks. They either try to find motifs that are over-represented in a given set of sequences, or they try to find motif arrangements that distinguish two given sets of sequences. Our method, in contrast, is intended for regression tasks. There are also several methods that learn models that characterize gene-expression responses in terms of sequence features. Some of these approaches first group expression values into discrete sets and then frame the problem as a classification task [2, 21]. Other methods use a two-phase approach that first identifies candidate motifs without the use of expression data, and then learns a regression model from expression data and these fixed sequence features [5, 23]. The key difference between our approach and these methods is that, in our approach, the real-valued response associated with each sequence is a training signal that has a direct influence on the sequence features represented by the model.

## 2 Approach

The task that we consider is to learn a function which maps a given discrete character sequence $\mathbf{x} = \{x_1, x_2, ..., x_L\}$ to a real-valued scalar $y$. In this section we describe the representation we employ and discuss the procedure we use for learning the models.

### 2.1 Representation

We assume that there are certain features, or motifs, present in each sequence $\mathbf{x}$ that determine the associated $y$ value. However, in the tasks that we consider, the learner is given only sequences and their response values, and must discover both the motifs and their locations in the sequences. Thus, our approach involves learning the structure and parameters of a hidden Markov model that represents these motifs. The other key component of our learned model is a regression function that maps from occurrences of the motifs to $y$ values. In particular, we associate certain states in the HMM with motifs, and represent the putative occurrences of motifs in a given sequence by keeping track of the number of times that each of these states is visited. That is, a subset of the states in the HMM $\{c_1, c_2, ..., c_N\}$ are designated as "counted" states, and a path through the model defines an integer vector $\mathbf{v} = \langle v_1, v_2, ...v_N \rangle$, where each $v_k$ is the number of visits to state $c_k$.

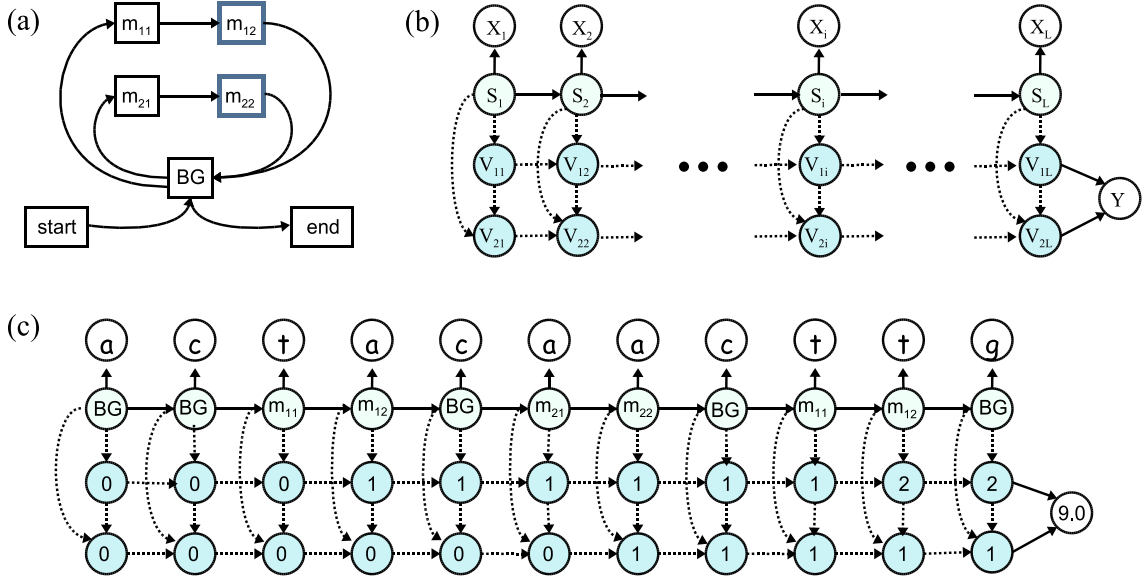

Figure 2: An HMM for regression and the corresponding graphical model. Panel **(a)** shows the state topology of a simple HMM that is able to represent occurrences of two types of motifs in given sequences. Each motif consists of exactly two DNA bases. For the $i$th motif, these bases are emitted by the states $m_{i1}$ and $m_{i2}$. The state labeled BG emits the remaining "background" of the sequence. To calculate the distribution over the possible motif occurrences for each sequence, we count visits to states $m_{12}$ and $m_{22}$. Panel **(b)** shows the structure of the corresponding graphical model when processing a sequence of length $L$. The $X_i$ variables represent the observable sequence characters. The $S_i$ variables represent the corresponding HMM state for each position in the input sequence. The $V_{1i}$ ($V_{2i}$) variables represent the number of visits to state $m_{12}$ ($m_{22}$) at or before the $i$th character in the input sequence. The $Y$ variable represents the real-valued response for the given instance. Probabilistic dependencies are illustrated using solid lines and deterministic dependencies are illustrated using dashed lines. Panel **(c)** shows the instantiation of variables in the graphical model for the instance (actacaacttg, 9.0) and a particular path through the HMM that visits $m_{12}$ twice and $m_{22}$ once.

More generally, we have uncertainty about the "correct" path through the model, and therefore uncertainty about the number of visits to each state $c_k$. Consider the HMM shown in Figure 2(a). There are two types of motifs, each two characters long. The motif occurrences are assumed to be interspersed with variable-length "background" sequence which is modeled by the BG state[1]. In this HMM, we count visits to each motif (*i.e.*, $c_1 = m_{12}$, $c_2 = m_{22}$). Figure 2(b) shows the corresponding graphical model when processing a sequence of length $L$. Each circle represents a random variable in the model, and edges represent direct dependencies. Probabilistic dependencies are shown with solid lines and deterministic dependencies are shown with dashed lines. Figure 2(c) shows the values taken on by the variables in the model for a case in which $\mathbf{x}$ = actacaacttg, $y = 9.0$, and we have assumed a particular path through the HMM. This path involves going through the top motif twice and the lower motif once. We discuss each of the random variables in turn.

[1] As an alternative to a self-transition, our background models include a probability distribution over lengths of subsequence, making these models *generalized* [4] or hidden *semi*-Markov models.

Each variable $X_i$ represents the $i$th character in the input sequence $\mathbf{x}$. The variable $S_i$ represents the HMM state that we are in after explaining the first $i$ characters of $\mathbf{x}$. This state depends directly on the previous state, and this dependency is encoded by the HMM transition parameters. The variable $X_i$ depends on the corresponding state variable $S_i$, and this relationship is encoded via the HMM emission parameters. In the problems we consider, these state sequences are hidden during both training and testing.

Each $V_{k,i}$ represents the number of visits to state $c_k$ in the paths through the HMM which explain the first $i$ characters of $\mathbf{x}$. These variables are also hidden and depend on the HMM state $S_i$ and the corresponding variable from the previous position, $V_{k,i-1}$. They are updated as follows:

$$P(V_{k,i} = v | s_i, V_{k,i-1}) = \begin{cases} P(V_{k,i-1} = v - 1) & \text{if } s_i = c_k \\ P(V_{k,i-1} = v) & \text{otherwise} \end{cases}$$
(1)

Moreover, as illustrated by the edges between the bottom two nodes in each column of Figure 2(b), we may represent dependencies among the $V_{k,i}$ variables at the $i$th position. Doing this enables us to model an arbitrary joint distribution characterizing the visits to the "counted" states.

Finally, the variable $Y$ in Figure 2(b) is the real-valued response associated with the sequence in question. Its value depends on the number of visits to all counted states after explaining the entire sequence. Thus, the last column of visit count variables in Figure 2(b) determines the response value, $y = f(\langle V_{1L}, ..., V_{NL} \rangle)$.

We represent $Y$ using a linear Gaussian model. Let $\mathbf{V}$ denote the vector of variables $\langle V_{1L}, ..., V_{NL} \rangle$, and let $\mathbf{v}$ denote a particular vector of visit counts $\langle v_{1L}, ..., v_{NL} \rangle$. Given a specific $\mathbf{v}$, this model represents the probability distribution of $Y$ as a Gaussian whose mean is a linear function of the visit-count vector:

$$p(Y|\mathbf{v}) \equiv N(\beta_1 v_{1L} + \beta_2 v_{2L} + ... + \beta_N v_{NL}, \sigma^2) \quad (2)$$

Here, each $\beta_k$ is an unknown model parameter which represents the contribution to the response variable of each occurrence of the motif represented by state $c_k$. The standard deviation $\sigma$ is also a model parameter to be learned.

Since the $\mathbf{V}$ variables are hidden, we may infer a distribution for $Y$ given a sequence $\mathbf{x}$, by marginalizing out $\mathbf{V}$, based on its likelihood given $\mathbf{x}$ which is computed by our hidden Markov model.

$$p(Y|\mathbf{x}) = \sum_{\mathbf{v}} p(Y|\mathbf{v}) P(\mathbf{v}|\mathbf{x}). \quad (3)$$

Note that (3) involves a summation over the possible values of $\mathbf{v}$. In general, the number of possible values can increase exponentially with the number of counted states $N$, but in practice, $N$ is a small number of features. There are tasks in which it is useful to have a larger number of counted states, and in many such cases, the calculations are still tractable because the HMM topology and the sequence characters in $\mathbf{x}$ prevent otherwise possible values of $\mathbf{v}$ from having a nonzero probability. If the number of possible values of $\mathbf{v}$ is still prohibitively large, then we may choose to consider only its most likely values, or we may instead calculate the expected value of $\mathbf{V}$ which is $\hat{\mathbf{v}} = \langle \hat{v}_1, \hat{v}_2, ..., \hat{v}_N \rangle$, where each $\hat{v}_k = \sum_\pi P(\delta(\pi)|\mathbf{x})$, where $\delta(\pi)$ deterministically maps a path $\pi$ through the HMM to the appropriate vector. $\hat{\mathbf{v}}$ is computed efficiently using dynamic programming. $p(Y|\mathbf{x})$ is then estimated as $p(Y|\hat{\mathbf{v}})$.

### 2.2 Parameter Learning

Given an HMM structure, we select parameter values to maximize the joint probability of the observed input sequences and their associated response values. Taking into account uncertainty in the "correct" path for each given sequence, and the dependencies represented in the model, we can express the objective function as:

$$\arg\max_{\Theta, \boldsymbol{\beta}, \sigma} \prod_{(\mathbf{x},y)} \sum_{\mathbf{s}} [P(\mathbf{s} : \Theta) P(\mathbf{x}|\mathbf{s} : \Theta) p(y|\mathbf{v} = \delta(\mathbf{s}) : \boldsymbol{\beta}, \sigma)] \quad (4)$$

where $s_i$ is the HMM state we are in at the $i$th character of $\mathbf{x}$. This product ranges over all of the $(\mathbf{x}, y)$ pairs in the training set, $\Theta$ represents the usual set of HMM state transition and character emission parameters, and $\boldsymbol{\beta}$ and $\sigma$ are the parameters of the regression model described above. Note that, because a given $\mathbf{s}$ maps deterministically onto a particular $\mathbf{v} = \delta(\mathbf{s})$, we do not need to sum over $\mathbf{v}$. We train using an expectation-maximization (EM) approach which is a slight modification of the standard Baum-Welch algorithm for HMMs [19].

**E-step:** EM algorithms are already widely used for training HMM models to represent sequence motifs (*e.g.* [1]). The difference between standard Baum-Welch and our approach is that we calculate the expected values for our hidden variables taking into account $y$ as well as $\mathbf{x}$. To accomplish this, we calculate a probability distribution over $\mathbf{V} = \langle V_{1L}, V_{2L}, ..., V_{NL} \rangle$ given $\mathbf{x}$, $y$ and our model parameters by considering each possible value for $\mathbf{v}$. The probability is given by

$$P(\mathbf{v}|\mathbf{x}, y : \Theta, \boldsymbol{\beta}, \sigma) = \frac{1}{Z} p(y|\mathbf{v}, \mathbf{x} : \boldsymbol{\beta}, \sigma) \quad (5)$$

where $Z$ is a normalization constant. This is the base-case initialization for the backward calculations involved in standard E-step of Baum-Welch. Apart from this, we compute the expected values for all hidden variables $V_{1,1}, V_{1,2}, ... V_{N,L-1}$ and $S_1, S_2, ..., S_L$ using the standard forward-backward calculations, which update the probability distribution over $V_{1L}, V_{2L}, ..., V_{NL}$ accounting for $\mathbf{x}$ and $\Theta$.

**M-step:** Apart from the initialization of the backward calculations described above, the estimation of our HMM parameters $\Theta$ are calculated using standard M-step of Baum-Welch. We choose our regression model parameters $\boldsymbol{\beta}$ and $\sigma$ using standard least-squares regression, except that the possible values for $\mathbf{v}$ given a training example $i$ are weighted by their likelihood given $\mathbf{x}_i$ and $y_i$. Thus, we minimize the total *expected* squared difference between the observed and predicted response values in the training set. This is calculated by marginalizing over the possible values for $\mathbf{v}$, according the their likelihood $P(\mathbf{v}|\mathbf{x}_i, y_i : \Theta, \boldsymbol{\beta}, \sigma)$, which is calculated in the E-step:

$$\hat{\boldsymbol{\beta}} = \arg\max_{\boldsymbol{\beta}} \sum_i^D \sum_{\mathbf{v}} P(\mathbf{v}|\mathbf{x}_i, y_i : \Theta, \boldsymbol{\beta}, \sigma)(y_i - \boldsymbol{\beta} \cdot \mathbf{v})^2 \quad (6)$$

where $D$ is the training set size. Let $H_k$ be the maximum number of visits to state $c_k$.[2] $\mathcal{V} = \prod_k^N H_k$ is the number of possible values of $\mathbf{v}$. Equation (6) has the closed-form solution:

$$\hat{\boldsymbol{\beta}} = (\mathbf{A}^T \boldsymbol{\Gamma} \mathbf{A})^{-1} \mathbf{A}^T \boldsymbol{\Gamma} \mathbf{b} \qquad (7)$$

where $\mathbf{A}$ is a $\mathcal{V}D \times N$ matrix of all possible values of $\mathbf{v}$ for each training example, $\mathbf{b}$ is a $\mathcal{V}D \times 1$ vector of the $y$ response values corresponding to each row of $\mathbf{A}$, and $\boldsymbol{\Gamma}$ is a $\mathcal{V}D \times \mathcal{V}D$ diagonal matrix, where each value $\gamma_i$ represents the likelihood of $\mathbf{v}$ in row $i$ of $\mathbf{A}$, given the appropriate training example, i.e. $\gamma_1 = P(\mathbf{v} = \langle 0, 0, ..., 0 \rangle | \mathbf{x_1}, y_1 : \boldsymbol{\Theta}, \boldsymbol{\beta}, \sigma)$.

$$\mathbf{A} = \begin{bmatrix} 0 & 0 & \cdots & 0 \\ & \vdots & & \\ H_1 & H_2 & \cdots & H_N \\ 0 & 0 & \cdots & 0 \\ & \vdots & & \\ H_1 & H_2 & \cdots & H_N \end{bmatrix}, \mathbf{b} = \begin{bmatrix} y_1 \\ \vdots \\ y_1 \\ y_2 \\ \vdots \\ y_D \end{bmatrix},$$

$$\boldsymbol{\Gamma} = \begin{bmatrix} \gamma_1 & 0 & 0 & \cdots & 0 \\ 0 & \gamma_2 & 0 & \cdots & 0 \\ \vdots & & \ddots & & \vdots \\ & & & \ddots & 0 \\ 0 & 0 & \cdots & 0 & \gamma_{\mathcal{V}D} \end{bmatrix}. \qquad (8)$$

The value for $\sigma$ is estimated from the (minimized) expected difference between our best fit line and the data points.

Again, if the number of possible values of $\mathbf{v}$ is prohibitively large, we can sample from the distribution of $\mathbf{v}$ or we can use the expected number of visits $\hat{v}_k$ to each $c_k$, and solve $\hat{\boldsymbol{\beta}} = (\mathbf{A}^T \mathbf{A})^{-1} \mathbf{A}^T \mathbf{b}$, where

$$\mathbf{A} = \begin{bmatrix} \hat{\mathbf{v}}_1 \\ \hat{\mathbf{v}}_2 \\ \vdots \\ \hat{\mathbf{v}}_D \end{bmatrix} \mathbf{b} = \begin{bmatrix} y_1 \\ y_2 \\ \vdots \\ y_D \end{bmatrix} \qquad (9)$$

and $\hat{\mathbf{v}}_i$ is the vector of expected visits calculated for the $i$th training sequence, $\mathbf{x}_i$.

### 2.3 Structure Learning

Our task includes learning the underlying model structure as well as parameters. This structure refers to the set of states and transitions that define the HMM topology. Although our regression approach applies to arbitrary HMM structures, we are primarily interested in the occurrence and arrangement of motifs. These motifs represent classes of short substrings with character preferences at each position. Whereas Figure 2(a) shows a model that can represent an arbitrary number of occurrences of two very short motifs, more sophisticated arrangements can be encoded in the HMM topology. Consider the structures shown in Figure 3. Here each shaded, rounded shape represents a sequence of states modeling a single motif. This model considers not only the presence of particular motifs, but also their logical arrangement. The counted states (although they may be visited at most once) correspond to each combination of these substrings, and thus the response variable $Y$ is a function of not only the presence of particular sequence characters, but it is also senstitive to whether or not they appear in the preferred order.

---

[2]In practice, to make our calculations more efficient, if there is a cycle in the HMM topology that includes state $c_k$, we set $H_k$ to the highest number of visits to state $c_k$ that can reasonably be expected to occur.

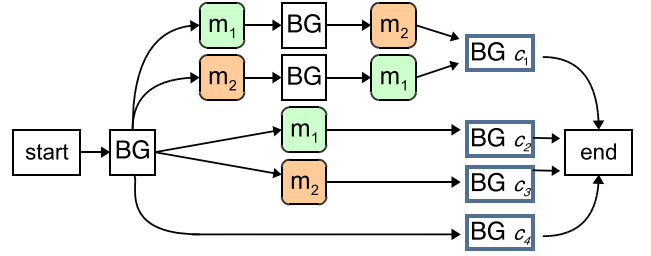

Figure 3: An HMM structure that considers both the presence and arrangement of over-represented sequence characters. This example has two such substrings, $m_1$ and $m_2$, which are chains of states but encapsulated here as single states. Other characters are explained by the BG background states. We count visits to states labeled $c_x$.

Instead of searching through the space of arbitrary HMM topologies by adding and removing individual states and transitions, our search operators are oriented toward the presence and arrangement of motifs. In our experiments, we begin our structure search with a single motif. We learn the parameters for this model and search HMM structure space by introducing additional motifs. We repeat this search process from different initial parameters some fixed number of times, and return the model structure and parameters that perform the best on the training set or a tuning set.

## 3 Results

Our task is to learn the structure and parameters of an HMM, as well as the parameters of our regression function. We hypothesize that an algorithm which uses the real-valued response associated with each input sequence to train HMM parameters is able to learn more accurate models than an approach which does not. To test this hypothesis, we compare our path-aggregate learning approach to a slightly less sophisticated two-phase version, where we first learn the HMM parameters $\Theta$ (using standard Baum-Welch), and then learn the parameters of our regression model $(\boldsymbol{\beta}, \sigma)$, from $\Theta$ and the observed sequence and response data. The key difference is that the regression model is just learned once in the two-phase approach, rather than iteratively refined as described in the previous section.

Given an input sequence $\mathbf{x}$, our models predict a probability density over the response $y$. In order to compare our method to the baseline method, we select a single value $\hat{y}$, by calculating the Viterbi (most likely) path through the HMM and then calculating the corresponding response according to the regression model, i.e. $\hat{y} = \boldsymbol{\beta} \cdot \mathbf{v}$, where $\mathbf{v}$ is the counted state visits implied by the Viterbi path.

To measure the accuracy of our models, we calculate the average absolute error on held-aside test sequences: error $= \frac{1}{T} \sum_i |y_i - \hat{y}_i|$. We test our learner on both simulated data and real gene expression data from yeast. For the yeast data, we believe the gene expression measurement is a function, in part, of a combination of short DNA motifs in the gene's promoter region, to which transcription factor proteins may bind to regulate the gene's expression. For the simulated data, we create such a situation by planting known motifs in simulated DNA sequence data.

For each simulated data experiment, We generate 200-character sequences from the alphabet {a,c,g,t}. We then plant 10-character motifs in each sequence. The number of times each motif is planted comes from a Poisson distribution with $\lambda = 1$. Only two of the motifs affect the response value, which is set to $-2 + 7 \times v_1 + 3 \times v_2 + \varepsilon$, where $v_i$ is the number of times motif $i$ was planted in the sequence, and $\varepsilon$ is random noise distributed normally from $N(0, 1)$. In our experiments, we vary the number of additional motifs (that do not affect response), and the "mutation rate," where a rate of $r$ means that $r$ characters in each motif are changed at random before the motif is planted.

The HMM model that we use is similar to the one shown in Figure 2(a), except that it varies in the number of motifs, and they are each 15 characters wide. We explore structures with one or two motifs, restarting 10 times with different initial settings. We keep the model with the highest accuracy on a held-aside tuning set. For each experiment, we generate 128 training sequences, 128 tuning sequences and 256 testing sequences, and we replicate each experiment several times.

Figure 4 shows how the accuracy of our learned models changes as a function of the mutation rate, and as a function of the number of additional planted motifs (apart from the two motifs that affect the response). The error rate using our approach is consistently less than that of the two-phase baseline, and tends to level off even as the number of mutations or additional motifs increases. Also, the recovery rate of the planted motifs is consistently higher using our integrated approach than that of the two-phase baseline. For instance, as we vary mutation rate and motif set size, we find that our approach returns the exact 10-character string of the motifs four times as often than the two-phase baseline. We conjecture that the reason our approach learns more accurate models than the two-phase baseline is because it is able to pick out the motifs that affect the response value instead of over-represented motifs which do not.

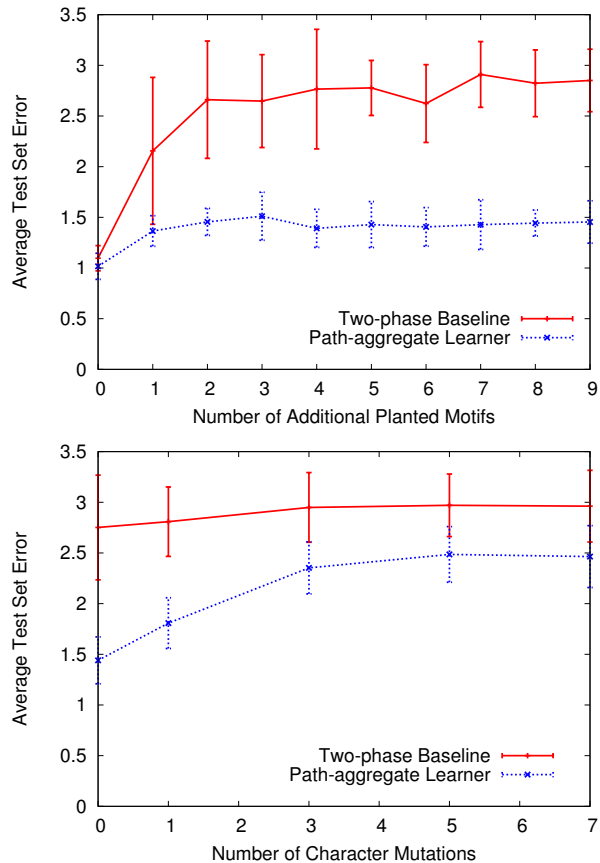

Figure 4: Test set error rate on simulated data comparing the path-aggregation learning approach to a two-phase baseline. Top: Test set error as a function of additional planted motifs that do not affect the response. Bottom: Test set error as a function of mutation rate (using five additional planted motifs).

To determine whether our approach can aid in the discovery of motifs in real genomic data, we use data from the yeast gene expression analysis of Gasch *et al.* (2000). In these experiments, yeast cells are put in a variety of stress conditions, such as heat shock or amino acid starvation, and measurements of gene expression are taken using microarrays to determine which genes' activity is increased or decreased specifically in these conditions. We choose 15 of these experiments that have the highest degree of differential expression and represent a variety of experimental conditions. From each of these, we select genes which are differentially expressed, and a control group with approximately the same number of genes. For each gene, we obtain 500 base pairs of promoter sequence from the University of California Santa Cruz genome browser [10].

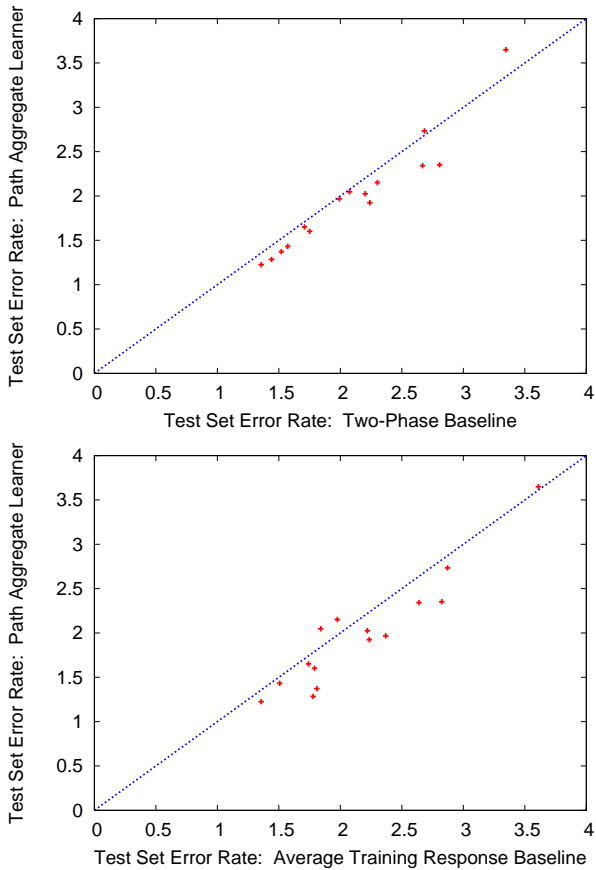

Figure 5: Test set error rate over 15 data sets, comparing our path-aggregate learner with the two-phase baseline (top) and the average training set response baseline (bottom). Each point represents one data set. Points are below the diagonal on datasets where our approach has a smaller error rate.

For these data sets, we use models similar to the ones we have previously shown to be well-suited to the task of identifying motifs in promoter data [18]. An example of the HMM structure is shown in Figure 3. These models are able to represent conjunctions of motifs occuring in specific orders. Instead of counting motif occurrences, the regression model considers which *combinations* of motifs occur in each sequence. We search over the space of possible structures by incrementally adding new motifs to the existing model. Each such addition affects several parts of the HMM topology. We limit this search to a maximum of two motifs, and we find in our experiments that both our approach and the baseline method return a variety of different HMM structures. Since the initial parameter values affect the results of EM training, the motif emission parameters are initialized by sampling from the training sequences.

As one additional baseline, we include a model that always predicts the average training set expression as the predicted response: $\hat{y} = \frac{1}{D}\sum_{i}^{D} y_i$. The results are shown in Figure 5. The top panel in the figure compares our approach to the two-phase basline. The bottom panel compares against the average-expression baseline. The models learned by our path-aggregate approach are more accurate than the two-phase baseline for 13 of the 15 data sets. Eight of these 13 are statistically significant at a $p$-value of 0.05, using a two-tailed, paired $t$-test over the ten cross-validation folds. Our models are more accurate than the training set average baseline for 12 of 15 data sets (10 of these are statistically significant).

## 4 Conclusion

We have presented a novel approach for learning HMM models for sequence-based regression tasks. Our approach involves simultaneously learning the structure and parameters of an HMM, along with a linear regression model that maps occurrences of sequence motifs to the response variable. Our experiments indicate that integrating the processes of learning the HMM and the associated regression model yields more accurate models than a two-phase baseline regression approach which first learns the HMM and then subsequently learns a regression model. We note that this baseline is fairly sophisticated, compared to many methods for sequence-based regression, in that it does not rely on a fixed, pre-defined set of features to represent each sequence being processed.

**Acknowledgements**

This research was supported in part by NIH grants T15 LM007359, R01 LM07050, and R01 GM077402. The authors would like to thank Audrey Gasch, Yue Pan and Tim Durfee for help with data and analysis.

## References

[1] T. Bailey and C. Elkan. Unsupervised learning of multiple motifs in biopolymers using expectation maximization. *Machine Learning*, 21:51–83, 1995.

[2] M. A. Beer and S. Tavazoie. Predicting gene expression from sequence. *Cell*, 117:185–198, 2004.

[3] Y. Bengio and P. Frasconi. An input output HMM architecture. In G. Tesauro, D. Touretzky, and T. Leen, editors, *Advances in Neural Information Processing Systems*, volume 7. MIT Press, Cambridge, MA, 1995.

[4] C. Burge and S. Karlin. Prediction of complete gene structures in human genomic DNA. *Journal of Molecular Biology*, 268:78–94, 1997.

[5] E. M. Conlon, X. S. Liu, J. D. Lieb, and J. S. Liu. Integrating regulatory motif discovery and genome-wide expression analysis. *Proc. of the National Academy of Sciences*, 100(6):3339, 2003.

[6] A. P. Gasch, P. T. Spellman, C. M. Kao, O. Carmel-Harel, M. B. Eisen, G. Storz, D. Botstein, and P. O. Brown. Genomic expression programs in the response of yeast cells to environmental changes. *Molecular Biology of the Cell*, 11(12):4241–57, 2000.

[7] T. Hastie, R. Tibshirani, and J. Friedman. *The Elements of Statistical Learning*. Springer-Verlag, 2001.

[8] T. Jaakkola, M. Diekhans, and D. Haussler. Using the Fisher kernel method to detect remote protein homologies. In *Proc. of the Seventh International Conf. on Intelligent Systems for Molecular Biology*. AAAI Press, 1999.

[9] A. Karali and I. Bratko. First order regression. *Machine Learning*, 26(2-3):147–176, 1997.

[10] D. Karolchik, A. Hinrichs, T. Furey, K. Roskin, C. Sugnet, D. Haussler, and W. Kent. The UCSC table browser data retrieval tool. *Nucleic Acids Research*, 32(1):D493–D496, 2004.

[11] S. Kramer. Structural regression trees. In *Proc. of the Thirteenth National Conf. on Artificial Intelligence*, pages 812–819. AAAI/MIT Press, 1996.

[12] A. Krogh, M. Brown, I. S. Mian, K. Sjolander, and D. Haussler. Hidden Markov models in computational biology: Applications in protein modeling. *Journal of Molecular Biology*, 238:54–61, 1994.

[13] J. Lafferty, A. McCallum, and F. Pereira. Conditional random fields: Probabilistic models for segmenting and labeling sequence data. In *Proc. of the Eighteenth International Conf. on Machine Learning*, pages 282–289. Morgan Kaufmann, 2001.

[14] C. Lawrence, S. Altschul, M. Boguski, J. Liu, A. Neuwald, and J. Wootton. Detecting subtle sequence signals: A Gibbs sampling strategy for multiple alignment. *Science*, 262:208–214, 1993.

[15] C. Leslie, E. Eskin, and W. Noble. The spectrum kernel: A string kernel for SVM protein classification. *Pacific Symposium on Biocomputing*, 7:566–575, 2002.

[16] N. Li and M. Tompa. Analysis of computational approaches for motif discovery. *Algorithms for Molecular Biology*, 1:8, 2006.

[17] C. Manning and H. Schütze. *Foundations of Statistical Natural Language Processing*. MIT Press, Cambridge MA, 1999.

[18] K. Noto and M. Craven. Learning probabilistic models of cis-regulatory modules that represent logical and spatial aspects. *Bioinformatics*, 23(2):e156–e162, 2007.

[19] L. R. Rabiner. A tutorial on hidden Markov models and selected applications in speech recognition. *Proc. of the IEEE*, 77(2):257–286, 1989.

[20] E. Segal and R. Sharan. A discriminative model for identifying spatial cis-regulatory modules. In *Proc. of the Eighth Annual International Conf. on Computational Molecular Biology (RECOMB)*, pages 141–149. ACM Press, 2004.

[21] Y. Yuan, L. Guo, L. Shen, and S. Liu. Predicting gene expression from sequence: A reexamination. *PLoS Comput Biol*, 3(11):e243, Nov 2007.

[22] Q. Zhou and W. H. Wong. CisModule: *De novo* discovery of cis-regulatory modules by hierarchical mixture modeling. *Proc. of the National Academy of Sciences*, 101(33):12114–12119, 2004.

[23] C. B. Z. Zilberstein, E. Eskin, and Z. Yakhini. Using expression data to discover RNA and DNA regulatory sequence motifs. In *Regulatory Genomics: RECOMB 2004 International Workshop*. Springer-Verlag, New York, NY, 2004.